\documentclass[runningheads]{llncs}

\usepackage{eccv}

\usepackage{eccvabbrv}

\usepackage{graphicx}
\usepackage{booktabs}
\usepackage[table]{xcolor} 

\usepackage[accsupp]{axessibility}  % Improves PDF readability for those with disabilities.

\usepackage{listings}
\usepackage{xcolor}

\usepackage{hyperref}

\usepackage{orcidlink}

\begin{document}

\title{SceneAssistant: A Visual Feedback Agent for Open-Vocabulary 3D Scene Generation}

\titlerunning{SceneAssistant}

% \author{Jun Luo\inst{1}\orcidlink{0000-1111-2222-3333} \and
% Jiaxiang Tang\inst{2}\orcidlink{1111-2222-3333-4444} \and
% Ruijie Lu\inst{1}\orcidlink{2222--3333-4444-5555} \and
% Gang Zeng\inst{1}\orcidlink{2222--3333-4444-5555}}
\author{Jun Luo\inst{1}\and
Jiaxiang Tang\inst{2}\and
Ruijie Lu\inst{1} \and
Gang Zeng\inst{1}}

\authorrunning{J. Luo \etal}

% Springer Heidelberg, Tiergartenstr.~17, 69121 Heidelberg, Germany

\institute{Peking University \and
NVIDIA}

\maketitle

\begin{figure}[tb]
  \centering
  \includegraphics[width=\linewidth]{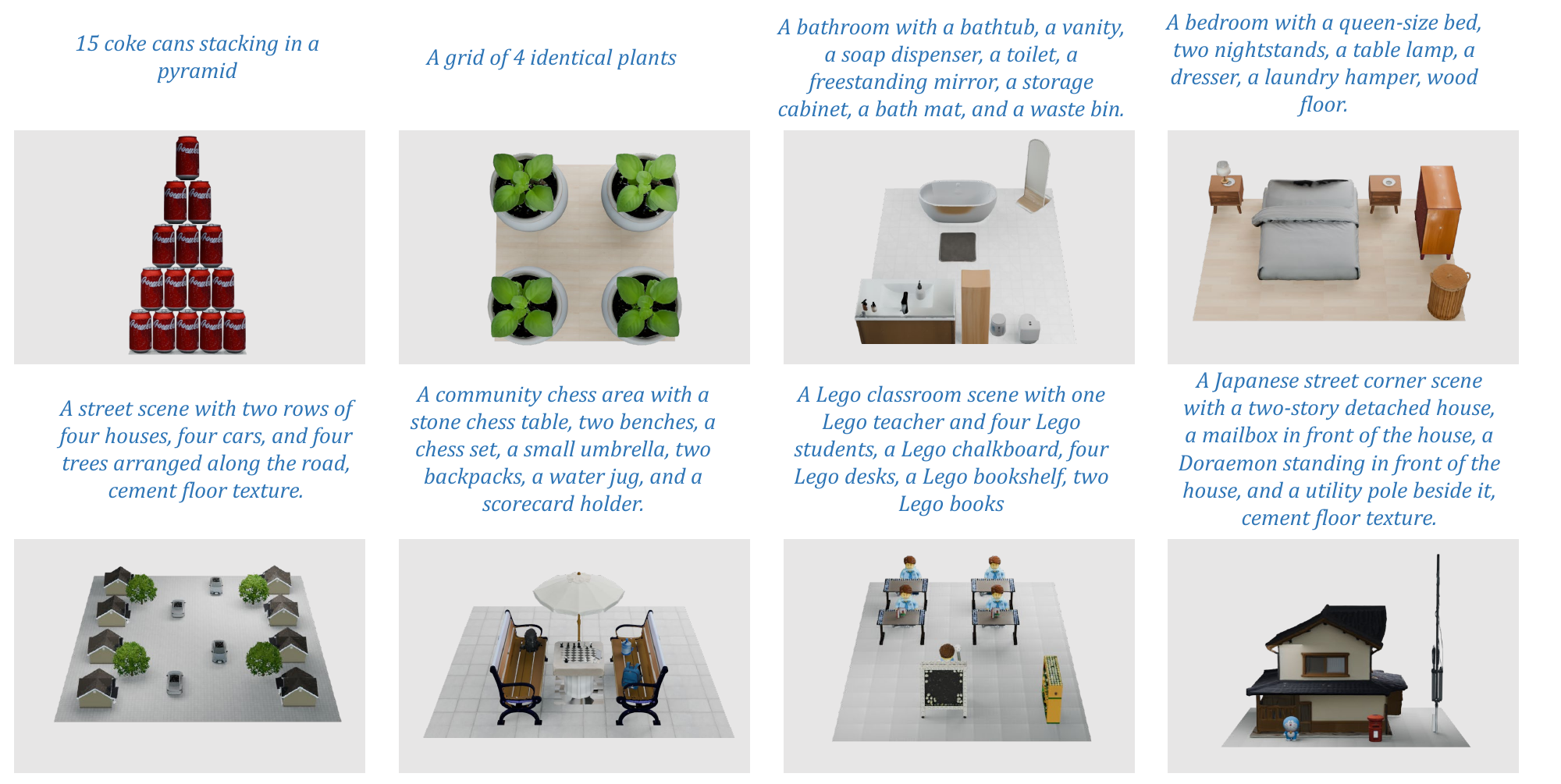}
  \caption{\emph{Open-vocabulary text-to-3D scene generation via SceneAssistant.} Each scene exhibits high fidelity to long-tail objects and intricate spatial constraints described in the text.}
  \label{teaserfig}
\end{figure}

\begin{abstract}
Text-to-3D scene generation from natural language is highly desirable for digital content creation. However, existing methods are largely domain-restricted or reliant on predefined spatial relationships, limiting their capacity for unconstrained, open-vocabulary 3D scene synthesis. In this paper, we introduce SceneAssistant, a visual-feedback-driven agent designed for open-vocabulary 3D scene generation. Our framework leverages modern 3D object generation model along with the spatial reasoning and planning capabilities of Vision-Language Models (VLMs). To enable open-vocabulary scene composition, we provide the VLMs with a comprehensive set of atomic operations (\eg, Scale, Rotate, FocusOn). At each interaction step, the VLM receives rendered visual feedback and takes actions accordingly, iteratively refining the scene to achieve more coherent spatial arrangements and better alignment with the input text. Experimental results demonstrate that our method can generate diverse, open-vocabulary, and high-quality 3D scenes. Both qualitative analysis and quantitative human evaluations demonstrate the superiority of our approach over existing methods. Furthermore, our method allows users to instruct the agent to edit existing scenes based on natural language commands.
%Our code will be publicly available to support future research.
Our code is available at \url{https://github.com/ROUJINN/SceneAssistant}
\end{abstract}

\section{Introduction}
\label{sec:intro}

3D scene generation\cite{3dscenesurvey} is a critical topic in computer vision and graphics, serving as a cornerstone for a wide range of applications, from creative industries such as filmmaking\cite{zhang2025generative} and gaming\cite{xu2024sketch2scene} to technical domains like robotics simulation and embodied AI research\cite{xia2026sage}. However, constructing 3D scenes typically requires significant manual effort and expertise in software, such as Blender. To address this, numerous research efforts have focused on automatic 3D scene generation from text\cite{holodeck,sceneweaver,sceneteller,reason3d,anyhome} or image\cite{tang2025towards,artiscene} inputs.

Many data-driven approaches \cite{li2024director3d, li2024dreamscene} generate scenes represented as NeRFs \cite{nerf} or 3D Gaussians \cite{3dgs}. While effective, these representations often lack editability, limiting their utility in downstream applications, and their generative capabilities are constrained by the diversity of available datasets \cite{3dfront}. Recent works \cite{holodeck, sun2025layoutvlm, sceneweaver} adopt retrieval-based methods that leverage large object datasets \cite{deitke2023objaverse, 3dfuture, 3dfront}, combined with the strong reasoning capabilities of Vision-Language Models (VLMs) \cite{gemini, gpt4}. In such pipelines, the VLM acts as a high-level planner that interprets input text and generates spatial constraints, which are subsequently resolved into object positions and orientations through search or optimization-based solvers.
However, a major limitation of these approaches is their reliance on a restricted set of predefined spatial relationship primitives (\eg, on, face to, in front of), which are often tailored to specific domains such as indoor environments. When user descriptions involve nuanced or complex configurations that fall outside this fixed vocabulary, the optimization process becomes poorly constrained or fails to capture the intended semantics. Consequently, these methods struggle to generalize to open-vocabulary scenes, often producing layouts that are suboptimal or even erroneous.

In this work, we tackle the challenge of open-vocabulary text-to-3D scene generation \cite{holodeck20, agentic3dspatiallycontextualizedvlms, scenelang, sceneweaver}. This objective implies that input text descriptions are not confined to specific indoor or outdoor domains but can encompass arbitrary environments and a diverse array of objects. Given the scarcity of large-scale and diverse 3D scene datasets, we propose to leverage the reasoning power of modern, rapidly advancing VLMs as the core controller for this task.
Contrary to the prevailing assumption that VLMs cannot directly infer spatial arrangements \cite{holodeck, sun2025layoutvlm} and thus necessitate external optimization procedures or predefined spatial primitives, we observe that modern VLMs \cite{wang2024exploring} already possess a latent degree of spatial awareness and planning proficiency. We argue that these inherent capabilities can be effectively elicited and harnessed through a carefully designed complete set of Action APIs. These APIs serve as a bridge, translating the VLM’s high-level spatial reasoning into concrete scene manipulations. By providing the VLM with a comprehensive suite of scene editing and viewpoint control tools (as detailed in \cref{actiontab}), we enable it to achieve a level of layout control comparable to that of human experts, without relying on rigid, predefined spatial templates.

Building upon these Action APIs, we develop an agentic framework following the ReAct \cite{yao2022react} paradigm, where the VLM acts as an autonomous agent that iteratively refines the scene. At each step, the agent performs reasoning, executes a specific action via our APIs, and receives visual feedback in the form of a rendered image of the current scene state. The integration of visual feedback not only provides rich environmental cues but also empowers the VLM to autonomously assess the quality of generated assets—provided by a state-of-the-art text-to-3D generator \cite{hunyuan3d}—and prune low-quality ones. This closed-loop mechanism effectively mitigates the inherent instability and stochasticity of 3D asset generators. This synergy between the iterative agentic framework and our expressive Action APIs allows the system to effectively bridge the gap between high-level semantic intent and precise spatial execution, enabling the synthesis of complex 3D scenes with unprecedented flexibility and structural coherence.
Furthermore, SceneAssistant supports an interactive human-agent collaborative workflow. By allowing users to intervene with real-time feedback or constructive requests, the system leverages VLM’s robust instruction-following capabilities to ensure strict alignment with user intent, effectively raising the system’s performance ceiling for complex scene synthesis.

As shown in \cref{teaserfig}, we conduct extensive experiments across a variety of scene types and descriptions to demonstrate the versatility of our method, including regular layouts, indoor scenes, outdoor scenes, and uncommon configurations. Both qualitative analysis and quantitative human evaluations show that our approach significantly outperforms baseline methods. We also provide ablation studies validating the design choices in our agentic framework.

In summary, our primary contributions are as follows:
\begin{itemize}
    \item We present a novel agentic framework for open-vocabulary 3D scene generation that operates through a pure visual-feedback-driven loop, effectively bridging the gap between high-level textual concepts and complex 3D scenes.
    \item We design a versatile suite of scene editing and viewpoint control APIs that elicits the latent spatial reasoning and planning capabilities of modern VLMs, enabling fine-grained scene manipulation without the need for predefined spatial relationships or external layout tools.
    \item Extensive experiments across diverse and unconventional scene descriptions demonstrate that our approach consistently outperforms existing state-of-the-art methods in terms of layout coherence, asset quality, and alignment with complex user intent.
\end{itemize}

We will release our code to facilitate future research in this direction.

\section{Related Works}

\subsection{3D Scene Synthesis}

Traditional 3D scene generation predominantly relies on data-driven layout distributions or procedural generation. Early works such as 3D-FRONT \cite{3dfront} provide high-quality indoor layout datasets, enabling research on autoregressive models (\eg, ATISS \cite{paschalidou2021atiss}) and diffusion models (\eg, DiffuScene \cite{tang2024diffuscene}). However, these approaches are typically confined to specific categories (\eg, bedrooms, living rooms) and struggle to accommodate diverse object types. An alternative paradigm is procedural generation, exemplified by Infinigen \cite{infinigen} and ProcTHOR \cite{ProcTHOR}, which construct large-scale scenes through physical rules and scripting. While SceneX \cite{zhou2025scenex} further enhances controllability in procedural generation, such methods often require intricate programming or parameterized templates. In contrast, SceneAssistant enables open-vocabulary scene construction via natural language and general-purpose Action APIs, eliminating the need for predefined layout templates.

\subsection{LLM-based Scene Planning}

With the emergence of large language models (LLMs) demonstrating strong reasoning capabilities, researchers have begun exploring their use as scene planners. LayoutGPT \cite{feng2023layoutgpt} leverages in-context learning to generate visual layouts, while SceneTeller \cite{sceneteller} translates language into 3D layout representations. To enhance operational precision, 3D-GPT \cite{3dgpt} and SceneCraft \cite{scenecraft} guide LLMs to directly write Blender Python code.  Holodeck \cite{holodeck}, LayoutVLM \cite{sun2025layoutvlm} and several other works \cite{holodeck20,ling2025scenethesis,liu2025worldcraft,Idesign} employ LLMs to generate spatial constraints and then search or optimize layouts. Reason3D \cite{reason3d} employs large reasoning models to directly specify the positions and orientations of objects. Although these works achieve text-to-scene mapping, they predominantly operate in an open-loop manner: generated layouts or code are executed without subsequent refinement based on rendering outcomes. SceneAssistant introduces a closed-loop visual feedback mechanism, enabling the model to iteratively adjust parameters by observing rendered images, akin to human creative workflows.

\subsection{Agentic Frameworks and Vision-Language Feedback}

Recently, agentic frameworks endowed with perception and decision-making capabilities have emerged at the forefront of 3D generation.
SceneWeaver \cite{sceneweaver} introduces a reflective agentic framework that unifies diverse synthesis methods through a standardized tool interface. Using a VLM-based planner, it iteratively refines the scene based on both quantitative physical metrics and qualitative visual reflections. To improve spatial reasoning, Liu \etal\cite{agentic3dspatiallycontextualizedvlms} incorporate dynamic geometry-aware contexts, such as scene hypergraphs, to manage complex inter-object relationships. TreeSearchGen\cite{treesearchgen} introduce a global-local tree search algorithm that enables VLMs to perform spatial reasoning and error correction through backtracking. Focusing on the requirements of embodied AI, SAGE \cite{xia2026sage} incorporates a physics-critic via a simulator-in-the-loop validation, ensuring that generated environments are suitable for robot training. SceneAssistant distinguishes itself by employing pure visual feedback: it directly leverages vision-language models (VLMs) to interpret rendered images and iteratively refines scenes using a complete set of Action APIs. This design closely mirrors the intuitive workflow of human creators in 3D environments, enabling open-vocabulary generation and fine-grained control without the need for specialized spatial data structures, external tools or pre-defined spatial relationships.

\section{Method}

\begin{figure}[tb]
  \centering
  \includegraphics[width=\linewidth]{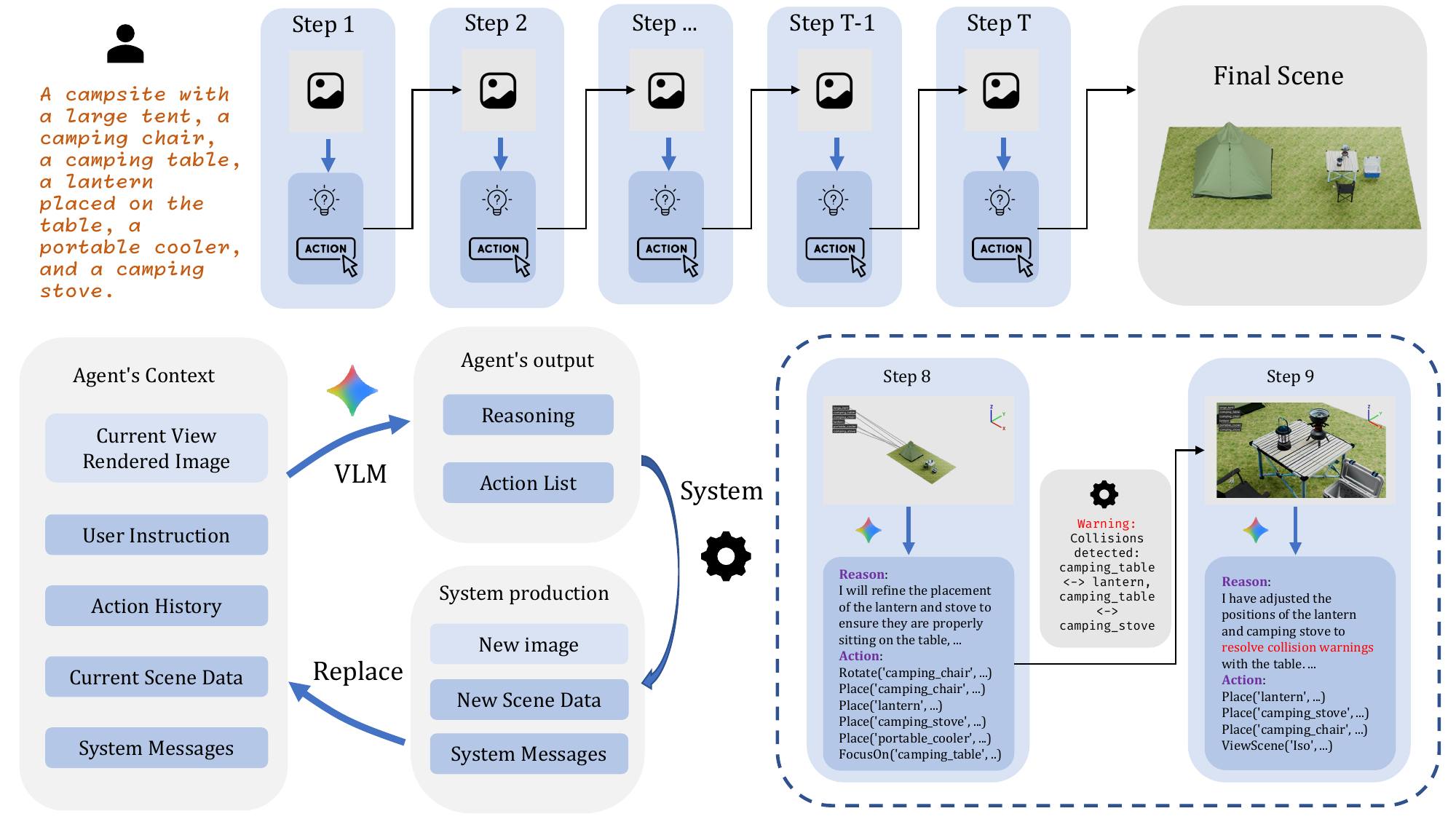}
  \caption{\emph{The SceneAssistant framework and its iterative scene generation process.} Our approach utilizes a vision-feedback-driven closed loop (bottom left) where a VLM agent processes multimodal context—including rendered views, scene metadata, and system messages—to generate reasoning and actions following the ReAct\cite{yao2022react} paradigm. This framework enables an iterative refinement process (top), progressively building complex scenes from step $1$ to $T$. As highlighted in steps $8$ and $9$ (bottom right), the agent can dynamically respond to system-generated collision warnings, performing corrective manipulations to resolve spatial inconsistencies and achieve a high-quality final 3D layout.}
  \label{methodfig}
\end{figure}

We begin by formalizing the task of open-vocabulary 3D scene generation. Given a user-provided natural language description $d$, the objective is to synthesize a 3D scene $s_d$ that faithfully aligns with the semantic and spatial constraints specified in $d$. In this context, the "open-vocabulary" nature implies that the description $d$ is agnostic to specific domain boundaries (\eg, restricted indoor or outdoor scenes) and may encompass an arbitrary array of objects. This necessitates a generation framework capable of handling a virtually unlimited set of categories and unconstrained spatial configurations beyond predefined templates.

To address these challenges, we introduce SceneAssistant, a vision-feedback-driven agentic framework specifically designed for open-vocabulary scene synthesis, as illustrated in \cref{methodfig}. The efficacy of SceneAssistant stems from two core design pillars: a comprehensive suite of Action APIs that abstracts complex 3D operations (\cref{sec:actionapi}), and an iterative agentic loop that leverages visual feedback for self-correction (\cref{sec:agenticframework}). Furthermore, our framework incorporates human-in-the-loop capabilities (\cref{sec:editing}), enabling users to perform natural-language-based scene editing.

\subsection{Complete Action APIs}
\label{sec:actionapi}

\begin{table}[tb]
  \caption{\emph{Comprehensive suite of Action APIs.} We define a functionally complete set of primitives for asset management, 6-DoF spatial manipulation, and visual perception control. These APIs abstract low-level engine operations into semantically intuitive commands, enabling the VLM agent to autonomously orchestrate 3D scene synthesis.}
  \label{actiontab}
  \centering
  \renewcommand{\arraystretch}{1.3} % Slightly increased for more breathing room
  
  % This enables alternating row colors starting from the 2nd row
  \rowcolors{2}{gray!10}{white} 
  
  \resizebox{\linewidth}{!}{%
  \begin{tabular}{@{}lll@{}}
    \toprule
    % Adds a soft blue background to the header row
    % \rowcolor{blue!15} 
    \textbf{Action Name} & \textbf{Arguments} & \textbf{Description} \\
    \midrule
    \texttt{Create}               & \texttt{name}, \texttt{description}               & Generate new object \\
    \texttt{Duplicate}            & \texttt{name}, \texttt{count}                     & Duplicate specified objects \\
    \texttt{Delete}               & \texttt{name}                                     & Delete objects \\
    \texttt{Translate}            & \texttt{name}, \texttt{axis}, \texttt{distance}   & Relative translation \\
    \texttt{Place}                & \texttt{name}, \texttt{position}                  & Absolute positioning \\
    \texttt{Rotate}               & \texttt{name}, \texttt{axis}, \texttt{angle\_degrees} & Set rotation angle (absolute) \\
    \texttt{Scale}                & \texttt{name}, \texttt{value}                     & Set scale (absolute, scalar or vector) \\
    \texttt{ViewScene}            & \texttt{view}, \texttt{zoom}                      & Reset camera to preset view \\
    \texttt{FocusOn}              & \texttt{target}, \texttt{view}, \texttt{zoom}     & Focus on specified object \\
    \texttt{RotateCamera}         & \texttt{horizontal}, \texttt{vertical}            & Relative camera rotation \\
    \texttt{MoveCamera}           & \texttt{direction}, \texttt{distance}             & Relative camera movement \\
    \texttt{GenerateFloorTexture} & \texttt{description}                              & Generate floor texture \\
    \texttt{Finish}               & ---                                               & Complete scene construction \\
  \bottomrule  
  \end{tabular}%
  }
\end{table}

We observe that modern VLMs, pre-trained on internet-scale datasets, already possess implicit latent spatial awareness\cite{ma2024spatialpin,vpsurvey} and reasoning proficiency \cite{wang2024exploring,guo2025deepseekr1}. Consequently, our design philosophy focuses on eliciting these inherent capabilities rather than imposing additional task-specific training or rules. We aim to keep the VLM within its optimal reasoning paradigm, focusing on high-level spatial planning rather than low-level implementation details. While we employ Blender as the underlying rendering engine, requiring a VLM to directly synthesize complex Blender Python scripts often introduces unnecessary syntactic overhead, which can compromise its strategic reasoning performance. To mitigate this, we abstract low-level operations into a set of simplified yet functionally complete Action APIs. This interface allows the agent to execute sophisticated scene manipulations through semantically intuitive commands. At each iteration, upon receiving visual feedback, the VLM selects a batch of actions from this suite to update the scene. The comprehensive list of actions is detailed in \cref{actiontab}, categorized into the following three functional groups.

\textbf{Object Addition and Deletion.} This category includes Create, Duplicate, and Delete. The Create operation is powered by a 3D asset generative model\cite{hunyuan3d}. Since the agent cannot know in advance what the generated object will look like based solely on the text description (\eg, whether a generated book is placed horizontally or vertically), the object is initially placed at the center of the scene immediately after creation. The agent can then observe the appearance of the created object in the rendered image of the next step and adjust its placement accordingly. To handle unsatisfactory generations from the 3D asset model, we provide the Delete action, allowing the agent to remove undesired objects and regenerate them with revised descriptions. This overall design ensures that the agent remains robust to the inherent uncertainties of 3D generative models.

\textbf{Object Manipulation.} We aim for our manipulation APIs to be complete, meaning that ideally any desired scene configuration can be achieved through a finite sequence of these actions. To meet this requirement, we provide the most direct method: Place, which sets an object's XYZ coordinates. Combined with Rotate, which sets the object's rotation angles along the XYZ axes, these two actions cover all six degrees of freedom for an object, enabling any arbitrary spatial arrangement. Furthermore, because objects generated by asset models are typically normalized in size, whereas real-world scenes often require varied object dimensions, we introduce the Scale action to give the agent control over object size. For fine-grained adjustments to an object's position within an existing scene, we also provide the Translate action as an additional refinement tool.

\textbf{Camera Control.} As the agent's actions are guided by visual feedback, it is crucial to ensure that this feedback is of high quality and provides sufficient information. This maximizes the effectiveness of the VLM's inherent general visual understanding and planning capabilities. To this end, we provide a complete set of camera APIs, where completeness means that ideally any camera state can be achieved through our provided actions. The two actions, RotateCamera and MoveCamera, fulfill this requirement. In most cases, however, the agent can rely on preset camera methods: ViewScene to observe the entire scene, which automatically frames all objects, and FocusOn, which allows the agent to examine a specific object and its immediate surroundings in detail.

A key merit of our framework is its modularity and inherent extensibility, allowing for the seamless integration of task-specific capabilities. For instance, we incorporate a GenerateFloorTexture action to facilitate the synthesis of context-aware ground textures, enhancing the visual coherence of the environment. To close the operational loop, the agent can take the Finish action once it autonomously verifies that the current scene state aligns with the user's objectives and has correct layout. Collectively, this integrated action space empowers the agent to navigate the vast complexity of open-vocabulary scene construction with high fidelity and autonomy.

\subsection{Agentic Framework with Pure Visual Feedback}
\label{sec:agenticframework}

Building upon the complete action APIs described in \cref{sec:actionapi}, we now detail the agent's workflow, as shown in bottom left of \cref{methodfig}. To prevent the agent from endlessly refining the scene without terminating, we set a maximum number of working steps, denoted as $T_M$. After $T$ steps, the agent outputs the final scene, caused by calling Finish or reaching the max step, so $T \le T_M$. At each step \(t\), the VLM receives a rendered image of the scene after the execution of the actions from step \(t-1\) (the camera position is determined by the camera actions invoked in previous steps). Furthermore, as the agent frequently needs to handle absolute coordinates and orientations of objects in the scene, we also provide it with the current data of all objects in the scene. Since the agent receives a new image at each step, providing all historical images as part of the dialogue would overload the agent with excessive information, especially over long horizons. Moreover, the relevance of earlier images to the current decision is limited. Therefore, we only supply the current image to the agent at each step. To inform the agent of the history, we provide it with the sequence of all actions executed in previous steps.

We follow the ReAct\cite{yao2022react} paradigm, requiring the agent to explicitly perform reasoning at each step before executing actions. We assume that the information provided to the agent should be highly reliable, whereas previous reasoning, memory, or plans may contain errors. The agent only needs to understand the current state and the target goal, and then iteratively optimize towards it. Consequently, we do not design or transmit previous reasoning, memory, or plans to the agent. After receiving all pertinent information, we instruct the agent to output a batch of actions, which enhances the efficiency of agent execution.

On the backend, to prevent objects from penetrating the ground plane, we automatically lift any object found below the ground to rest upon it. Preventing objects from floating in mid-air, however, relies on the agent's inherent visual capabilities. We also incorporate a system messages mechanism. If issues arise after the execution of actions in the current step, they are reported to the agent via system messages in the subsequent step. The application scenarios for system messages include the following: (1) We employ a BVH tree-based algorithm\cite{bvhtree} to automatically detect collisions among meshes in the scene. If a collision is detected, a corresponding notification is added to the system message for the agent. (2) We impose constraints on action sequences: object creation APIs must be executed in a separate batch and cannot be interleaved with object manipulation actions. This ensures the agent clearly observes the generated object's appearance before attempting to place it. Violation of this principle results in the rejection of the entire action batch, and the agent is informed via a system message. (3) Users can add edit instruction to SceneAssistant's execution trajectory, which are then passed to the VLM via system messages. Detailed explanation is available in \cref{sec:editing}.

To facilitate robust spatial reasoning, we employ a visual prompting mechanism\cite{treesearchgen,ma2024spatialpin,vpsurvey,chen2023understandingvp}, as exemplified in steps 8 and 9 of \cref{methodfig}. Specifically, each object in the rendered view is annotated with a semantic label corresponding to its unique name, enabling the VLM to distinguish between different instances and establish precise object grounding. Additionally, a coordinate axis HUD (Heads-Up Display) is overlaid on the viewport to provide a persistent global reference frame. This visual anchor assists the agent in mapping 2D perspective cues into 3D action parameters, effectively bridging the gap between screen-space observation and 6-DoF scene manipulation.

\subsection{Scene Editing Capability}
\label{sec:editing}

\begin{figure}[tb]
  \centering
  \includegraphics[width=\linewidth]{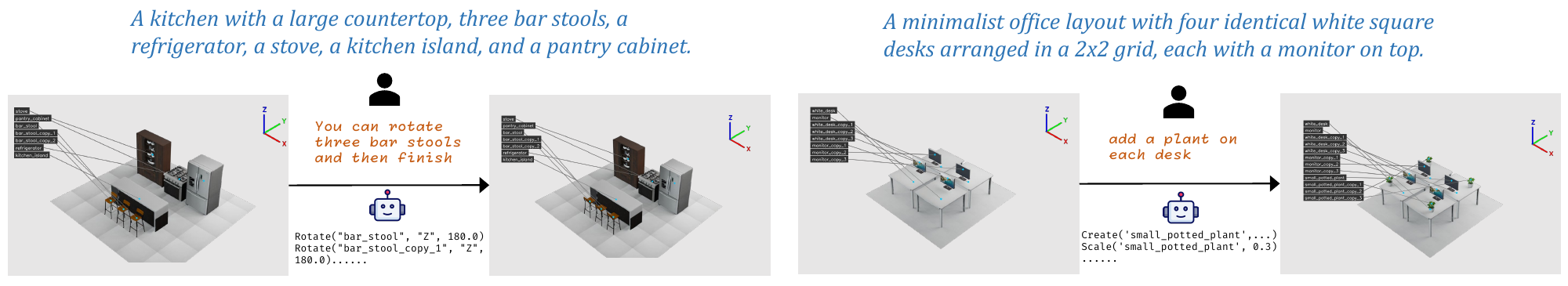}
  \caption{\emph{Human-agent collaboration for interactive scene editing.} SceneAssistant effectively interprets user-provided message and produces the corresponding scene.}
  \label{editfig}
\end{figure}

A key observation in modern VLMs is the performance discrepancy between their robust instruction-following capabilities and their occasionally limited fine-grained visual perception\cite{zhang2025scaling}. To capitalize on the former while mitigating the latter, SceneAssistant supports an interactive human-agent collaborative workflow. Users can intervene at any point of the agent's execution trajectory by injecting specific requirements via system messages. These requirements can range from corrective feedback on existing layouts to constructive requests for scene expansion. As illustrated in \cref{editfig}, the editing capability is often utilized for final-stage refinements or complex density adjustments (\eg, adding plants to every desk). After the agent reaches the maximum step limit $T$ and produces a near-complete layout, a single round of human-guided feedback is typically sufficient to rectify minor inaccuracies or finalize the scene composition, ensuring the output achieves the desired aesthetic and functional quality. The editing mechanism ensures the generated scene strictly aligns with the user's intent and significantly raises the system's performance ceiling.

\section{Experiments}

\subsection{Implementation Details}
Our framework utilizes Gemini-3.0-Flash \cite{gemini} as the core VLM backbone. For 3D asset generation, we employ a multi-stage pipeline centered on Hunyuan3D \cite{hunyuan3d}. Since Hunyuan3D is an image-conditioned generative model, we first synthesize high-quality object images using Z-Image \cite{zimage}. These images undergo a background removal process to obtain clean foreground masks before being fed into Hunyuan3D for 3D mesh reconstruction. The final scene assembly and high-fidelity rendering are performed using the Blender engine. We set max operating steps $T_M$ of SceneAssistant as 20. To prevent SceneAssistant from endlessly adding objects to refine the scene, we explicitly specify the existing objects in the scene prompt to eliminate semantic ambiguity. To showcase the autonomous capacity of SceneAssistant and ensure a fair comparison, all experimental results are generated without human intervention unless specified. This underscores the system’s capacity for independent synthesis, even though human-agent collaboration is supported for fine-grained editing.

% Crucially, to demonstrate the inherent spatial reasoning and execution capabilities of our framework, all scenes presented in the experimental section are generated fully autonomously by the agentic loop without any human intervention, unless explicitly stated otherwise. While SceneAssistant supports human-agent collaboration as a supplementary feature for nuanced editing, the results shown here testify to the system's capacity for independent and unconstrained scene synthesis.

\begin{figure}[tb]
  \centering
  \includegraphics[width=\linewidth]{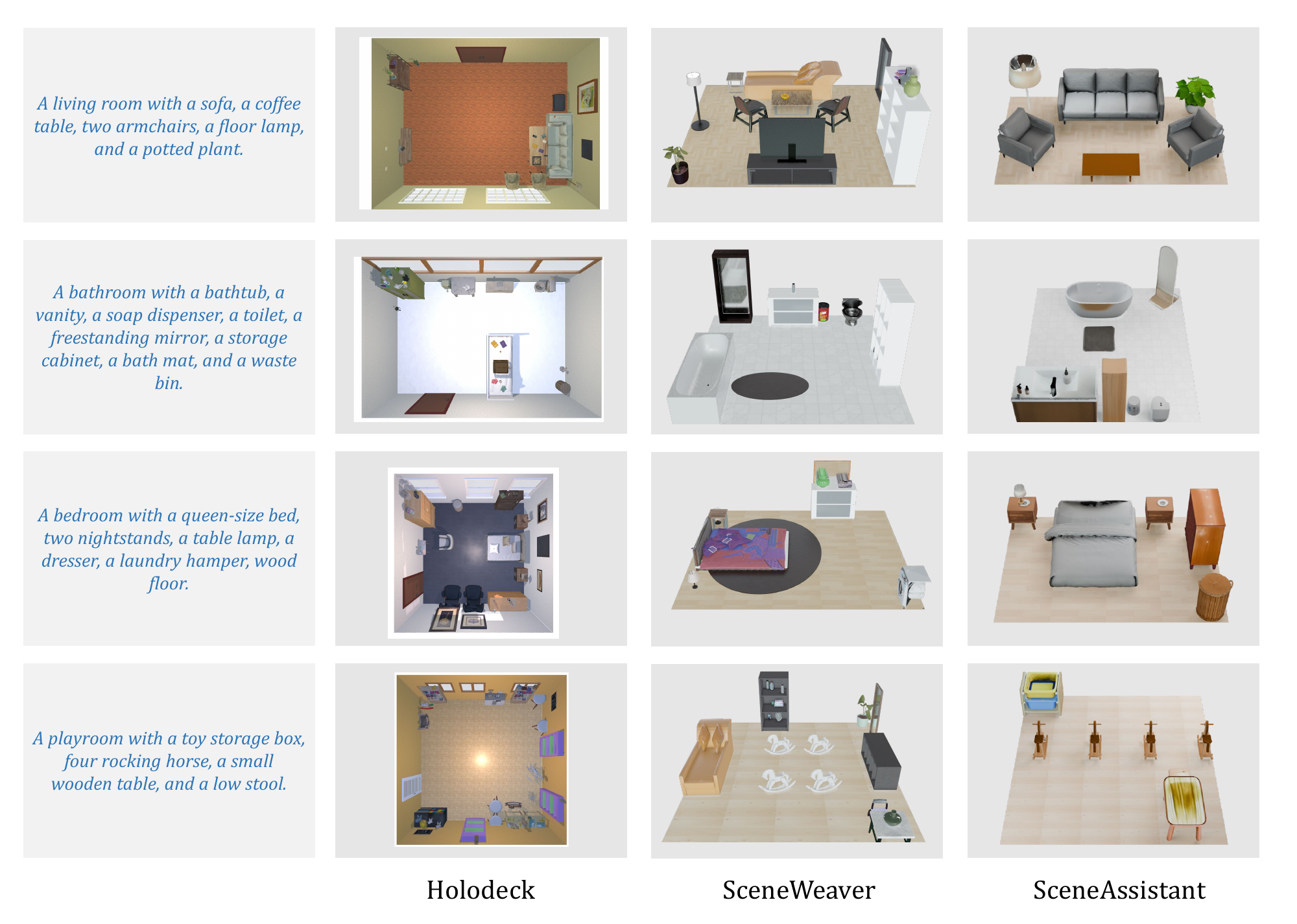}
  \caption{\emph{Qualitative comparison of indoor scene generation.} We compare SceneAssistant with Holodeck\cite{holodeck} and SceneWeaver\cite{sceneweaver} across various indoor categories. SceneAssistant performs precise spatial arrangements and faithfully reconstructs nuanced objects often simplified or omitted by existing baselines.}
  \label{indoorfig}
\end{figure}

\subsection{Indoor Scene Generation}

\subsubsection{Baselines}
We evaluate our method against two representative state-of-the-art approaches: Holodeck \cite{holodeck} and SceneWeaver \cite{sceneweaver}. To ensure a rigorous and fair comparison, we adapted the official implementation of SceneWeaver to align its underlying infrastructure with our setup. Specifically, we replaced its original VLM backbone (GPT-4 \cite{gpt4}) with Gemini-3.0-Flash to maintain consistency in reasoning capabilities across methods. Furthermore, we unified the floor texture assets used in SceneWeaver with those in SceneAssistant. Regarding the rendering engines, both SceneAssistant and the adapted SceneWeaver utilize Blender EEVEE, whereas Holodeck relies on Unity’s built-in assets and rendering pipeline. Both baselines utilize the Objaverse \cite{deitke2023objaverse,objaversexl} dataset for object retrieval.

\subsubsection{Results}

As shown in \cref{indoorfig}, our method achieves performance comparable to or even better than previous approaches specifically designed for indoor scene synthesis. Holodeck and SceneWeaver define spatial relationships among objects (\eg, ``against\_wall'') to determine their relative positions, which often results in objects being placed flush against walls. In contrast, our method directly predicts absolute coordinates, allowing for more flexible and diverse arrangements without being constrained by predefined spatial relations. Moreover, our approach better aligns with diverse user descriptions and does not introduce extraneous objects (\eg, cabinets or windows) by default. Overall, although our method is not explicitly designed for indoor scenes, it demonstrates competitive or superior performance in various indoor settings compared to specialized approaches.

\subsection{Open-Vocabulary Scene Generation}
\label{sec:ovexperiment}

\subsubsection{Baselines} 
As existing indoor scene generation methods typically lack support for open-vocabulary text-to-3D synthesis, and to the best of our knowledge, no publicly available works provide comparable functionality, we develop two variants of our framework to serve as baselines:
(1) \textbf{NoActionAPI}: This variant employs the same agentic framework as SceneAssistant but lacks specialized Action APIs. Instead, the agent is required to output the entire scene configuration (including object placement, rotation, camera parameters, and asset descriptions) through a comprehensive JSON structure. While the JSON representation is theoretically as expressive as our Action APIs, it compels the agent to manage low-level data structures manually.
(2) \textbf{NoVisFeedback}: The VLM performs one-shot generation without any subsequent visual feedback. To ensure a fair comparison regarding object knowledge, we provide the VLM with a grid-view of pre-generated 3D assets (the ones generated by SceneAssistant) to inform it of the objects' scale and orientation before it generates the final scene layout via JSON.

\begin{figure}[tb]
  \centering
  \includegraphics[width=\linewidth]{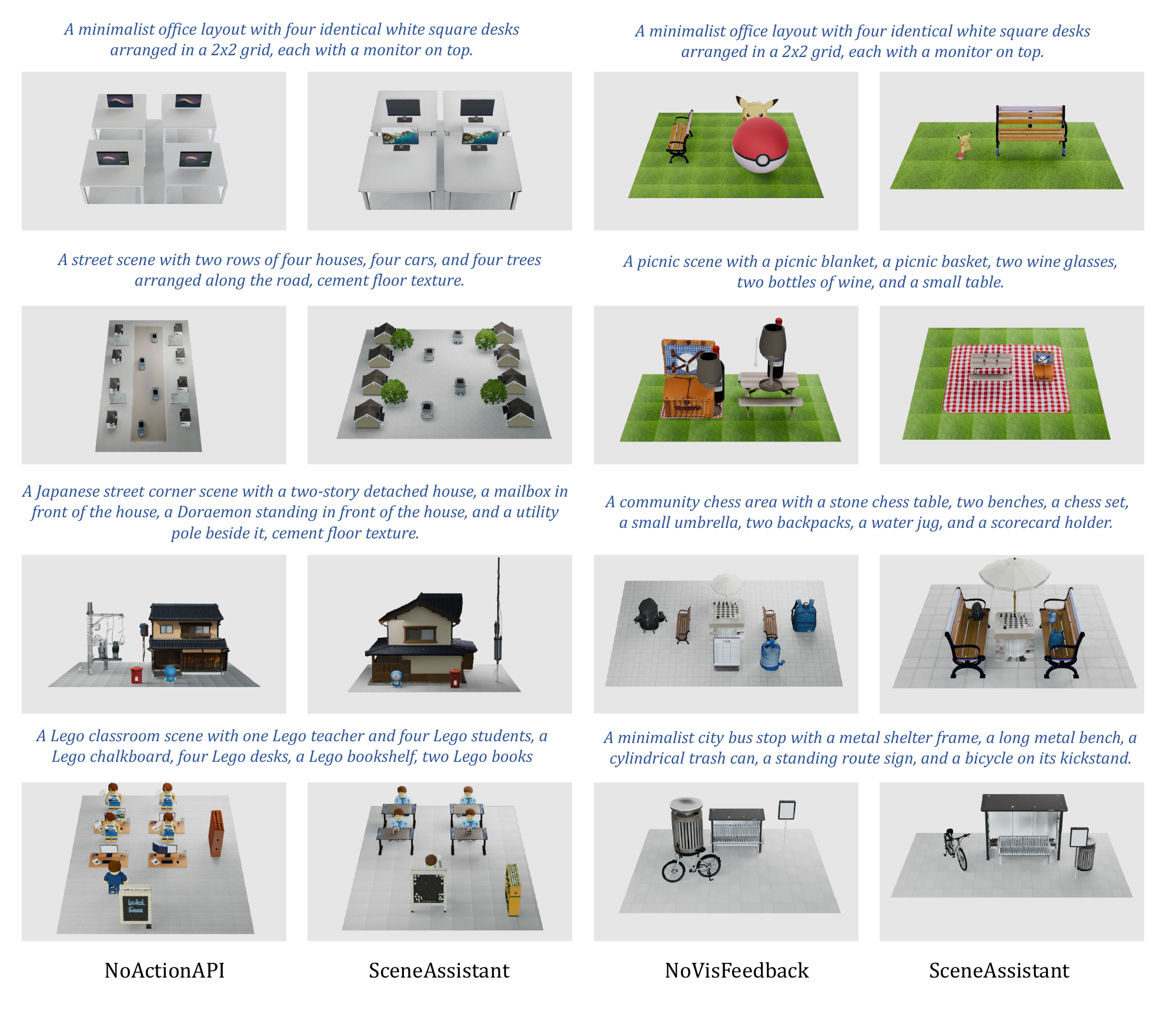}
  \caption{\emph{Qualitative results of open-vocabulary scene generation.} SceneAssistant consistently produces superior spatial layouts and semantic consistency.}
  \label{openvocabfig}
\end{figure}

\subsubsection{Results} 
As illustrated in \cref{openvocabfig}, SceneAssistant significantly outperforms both baselines. Although NoActionAPI operates with the same degrees of freedom and visual feedback, its performance is hindered by the complexity of direct JSON manipulation, which results in issues such as incorrect object orientation and unaddressed failures in 3D object generation. We attribute this to the cognitive distraction caused by forcing the agent to generate entire JSON strings, where the model fails to concentrate on fine-grained spatial adjustments. In contrast, our Action APIs abstract away the underlying complexity, allowing the agent to focus on high-level spatial reasoning and iterative improvement. Furthermore, the NoVisFeedback baseline struggles with relative scaling and object orientation, as the lack of a closed-loop feedback mechanism prevents the model from perceiving and correcting spatial misalignments in the integrated scene. The substantial performance gap between our full framework and these baselines underscores the necessity of both structured Action APIs and continuous visual grounding for high-quality open-vocabulary scene synthesis.

\subsection{Quantitative Evaluation}

\begin{table}[tb]
  \caption{\emph{Quantitative comparison across different scene categories}. The upper section reports the average scores for 8 indoor scenes, comparing SceneAssistant against state-of-the-art baselines. The lower section presents results for 22 open-vocabulary scenes under various configurations. Layout and Object Quality are rated on a 1--10 scale.}
  \label{quantitab}
  \centering
  \footnotesize
  % 使用 tabular*，宽度设为 \linewidth
  % @{\extracolsep{\fill}} 是关键，它会自动把多余的空间填入列间距
  \begin{tabular*}{0.98\linewidth}{@{\extracolsep{\fill}}lccc@{}}
    \toprule
    Method & Layout Correctness$\uparrow$ & Object Quality$\uparrow$ & Preference$\uparrow$ \\
    \midrule
    Holodeck\cite{holodeck} & 4.475 & 4.763 & 6.25\% \\
    SceneWeaver\cite{sceneweaver} & 5.800 & 6.150 & 36.25\% \\
    SceneAssistant & \textbf{6.888} & \textbf{6.950} & \textbf{61.25\%} \\
    \midrule
    NoActionAPI & 7.005 & 6.591 & 35.91\% \\
    NoVisFeedback & 6.255 & 5.673 & 26.82\% \\
    SceneAssistant & \textbf{7.600} & \textbf{7.277} & \textbf{65.00\%} \\
    
    \bottomrule
  \end{tabular*}
\end{table}

We conducted a rigorous quantitative evaluation across 30 diverse test cases, which include 8 curated indoor scenes and 22 unconstrained open-vocabulary scenarios. To ensure statistical robustness, each generated scene was independently assessed by 10 human participants. Given that automated metrics can be misleading: a low collision rate or a high number of objects, for instance, can not imply whether the scene semantically matches the text or appeals to human observers, we adopt a subjective evaluation protocol. Participants are required to rate each generated scene on a 1–10 scale along two primary dimensions: \textbf{Spatial Layout Correctness}, which measures whether the positioning and orientation of objects adhere to common-sense spatial reasoning and the given textual instructions; and \textbf{Object Quality}, which evaluates the relative object scales and the structural integrity of the generated assets. In addition, we report the \textbf{Human Preference Rate}, defined as the percentage of times a given method's output was selected as a preferred result (multiple selections were allowed). As summarized in \cref{quantitab}, SceneAssistant consistently outperforms all baselines across both indoor and open-vocabulary environment generation tasks, demonstrating superior alignment with human perference and textual instructions.

\subsection{Ablation Study}

% \begin{figure}[tb]
%   \centering
%   \includegraphics[width=\linewidth]{figs/novisualpromptfig.pdf}
%   \caption{\emph{Ablation of visual prompting mechanism}. Compared to our full SceneAssistant, the NoVisualPrompt variant lacks precise spatial grounding, leading to suboptimal object placement and a failure to accurately reflect the complexity of the input description.}
%   \label{novisualpromptfig}
% \end{figure}

% \begin{figure}[tb]
%   \centering
%   \includegraphics[width=\linewidth]{figs/nocollisionfig.pdf}
%   \caption{\emph{Ablation of collision detection mechanism}. Red circles highlight instances of interpenetration. Without explicit collision feedback, the agent fails to resolve geometric conflicts, demonstrating the necessity of collision check for ensuring 3D physical plausibility.}
%   \label{nocollisionfig}
% \end{figure}

\begin{figure}[tb]
  \centering
  \includegraphics[width=\linewidth]{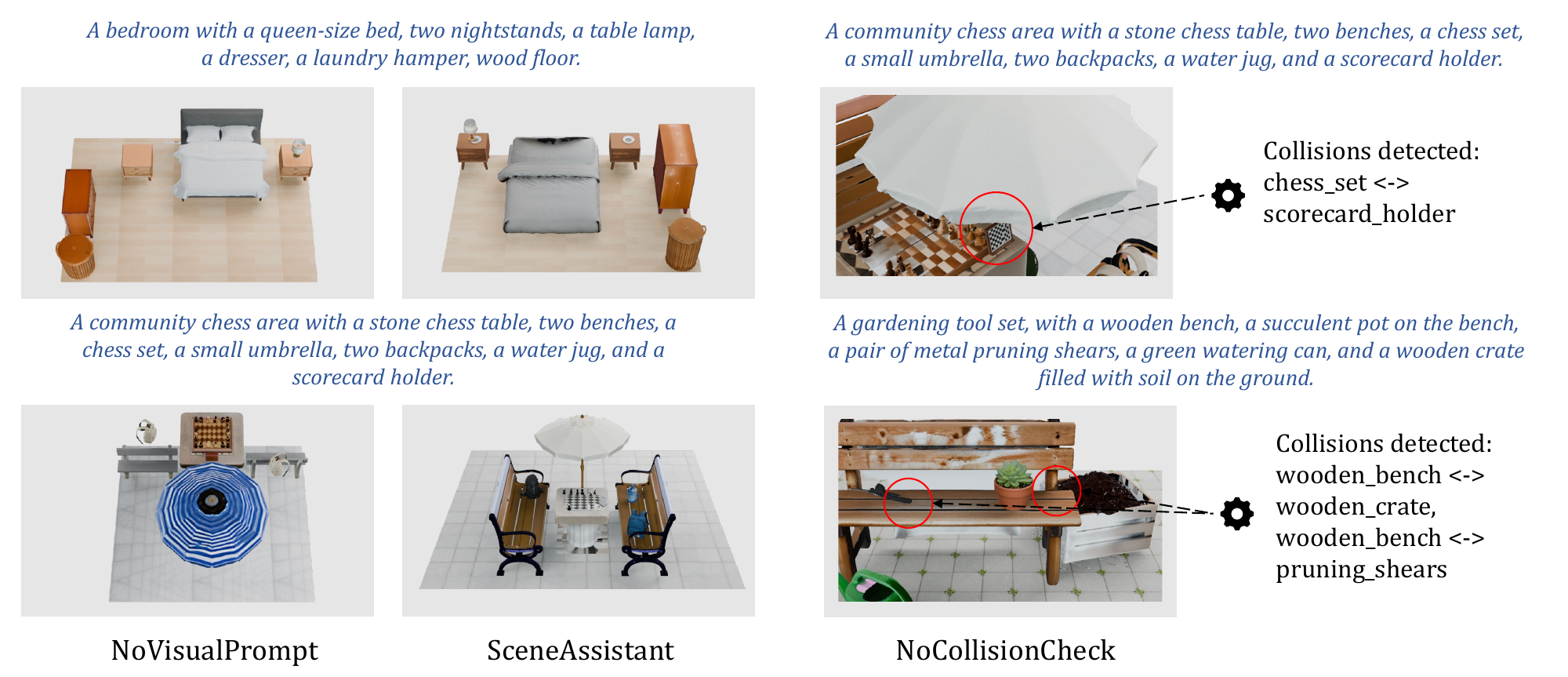}
  \caption{\emph{Ablation study.} The left side compares SceneAssistant with NoVisualPrompt variant, where the absence of visual grounding leads to disorganized layouts. The right side illustrates the NoCollisionCheck variant, with red circles highlighting physical interpenetrations that the agent fails to resolve without explicit feedback.}
  \label{ablationfig}
\end{figure}

To further validate the efficacy of our agentic framework and the contribution of individual components, we conduct extensive ablation studies. While the baselines in \cref{sec:ovexperiment} already reflect the impact of the action APIs and visual feedback, we focus here on two design choices: (1) \textbf{No Visual Prompting:} We remove all visual overlays, such as object bounding box labels and the coordinate axis HUD, from the rendered images provided to the agent. This test evaluates whether the agent can maintain spatial awareness and object-specific control without explicit visual prompting. (2) \textbf{No Collision Check:} We omit the explicit collision detection feedback from the system prompt. In this setting, the agent must rely solely on raw visual feedback to implicitly infer and avoid physical interpenetrations.

Qualitative comparisons are illustrated in \cref{ablationfig}. As shown in the left portion of the figure, removing visual prompts significantly degrades the agent's ability to precisely localize and manipulate objects, often resulting in disorganized layouts or missing components compared to the full SceneAssistant. This underscores the importance of visual-semantic alignment for fine-grained scene synthesis. Furthermore, the right portion of \cref{ablationfig} demonstrates that relying on implicit visual feedback alone is insufficient for maintaining physical plausibility. These results confirm that our integrated visual prompting and collision-checking mechanisms are essential for generating both visually coherent and physically realistic 3D scenes.

\section{Conclusion}
In this paper, we presented SceneAssistant, a novel agentic framework for open-vocabulary 3D scene generation. By shifting the paradigm from rigid spatial primitives or external layout solvers to a visual-feedback-driven closed loop, SceneAssistant effectively bridges the gap between high-level natural language instructions and complex 3D spatial reasoning. Central to our framework is a comprehensive suite of Action APIs, which provides the essential operational primitives for the VLM to execute precise 3D spatial planning. Our experimental results, spanning both standard indoor environments and unconventional open-domain scenarios, show that SceneAssistant consistently outperforms existing state-of-the-art methods in terms of spatial coherence, asset fidelity, and adherence to complex user intent. Furthermore, the framework's support for human-agent collaboration provides a flexible interface for creative 3D content creation. We believe that SceneAssistant serves as a significant step towards more autonomous and versatile 3D world modeling, with broad implications for downstream fields.

\bibliographystyle{splncs04}
\bibliography{main}

\clearpage % 确保另起一页

\begin{center}
% \Large 和 \large 控制字体大小，可以根据你的模板微调
\Large \textbf{Supplementary Material For SceneAssistant: A Visual Feedback Agent for Open-Vocabulary 3D Scene Generation}
\end{center}

\appendix

\setcounter{figure}{0}
\renewcommand{\thefigure}{S\arabic{figure}}
\setcounter{table}{0}
\renewcommand{\thetable}{S\arabic{table}}
\setcounter{equation}{0}
\renewcommand{\theequation}{S\arabic{equation}}

\section{More Implement Details}

\subsection{Gallery Results}

\begin{figure}[tb]
  \centering
  \includegraphics[width=\linewidth]{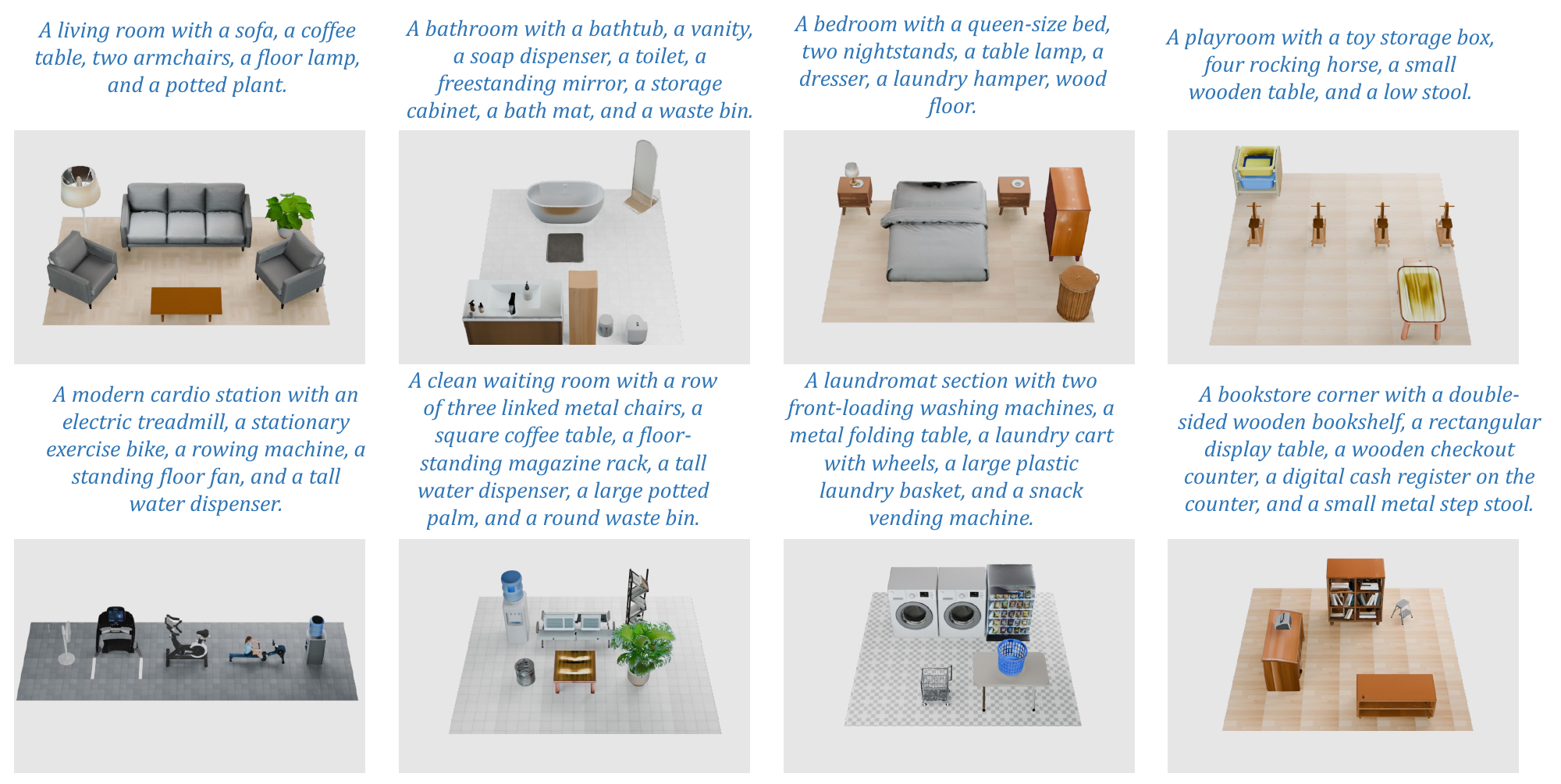}
  \caption{Gallery of rendered images for scenes generated by SceneAssistant in indoor scenarios.}
  \label{gallery_indoor}
\end{figure}

\begin{figure}[tb]
  \centering
  \includegraphics[height=0.9\textheight]{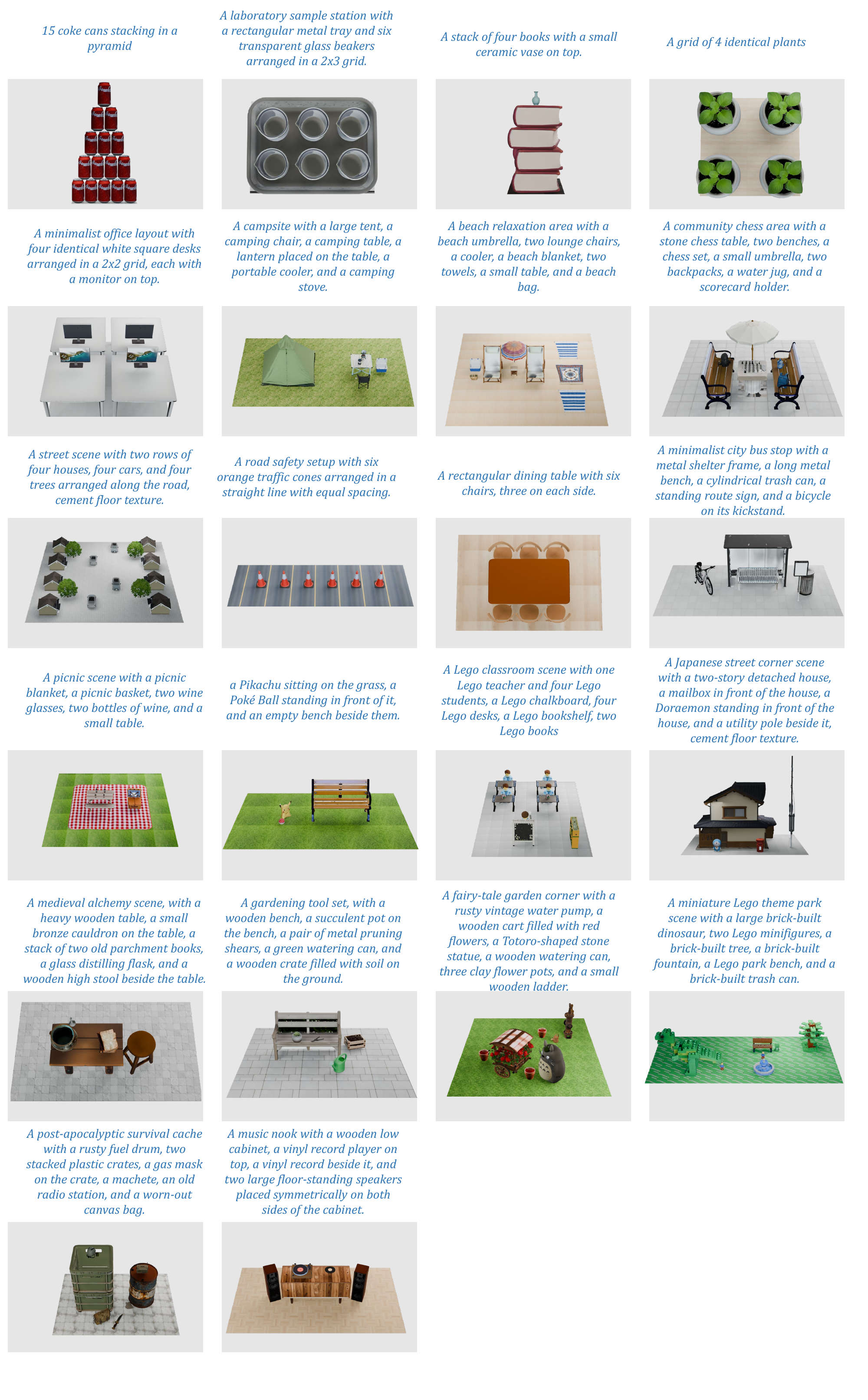}
  \caption{Gallery of rendered images for scenes generated by SceneAssistant in open-vocabulary scenarios.}
  \label{gallery_open}
\end{figure}

We provide rendered images of all generated scenes (8 indoor scenes and 22 oepn-vocabulary scenes) produced by SceneAssistant under both indoor and open-vocabulary scenarios, as shown in \cref{gallery_indoor} and \cref{gallery_open}.

\subsection{Detailed Workflow of SceneAssistant}

\begin{figure}[tb]
  \centering
  \includegraphics[height=0.9\textheight]{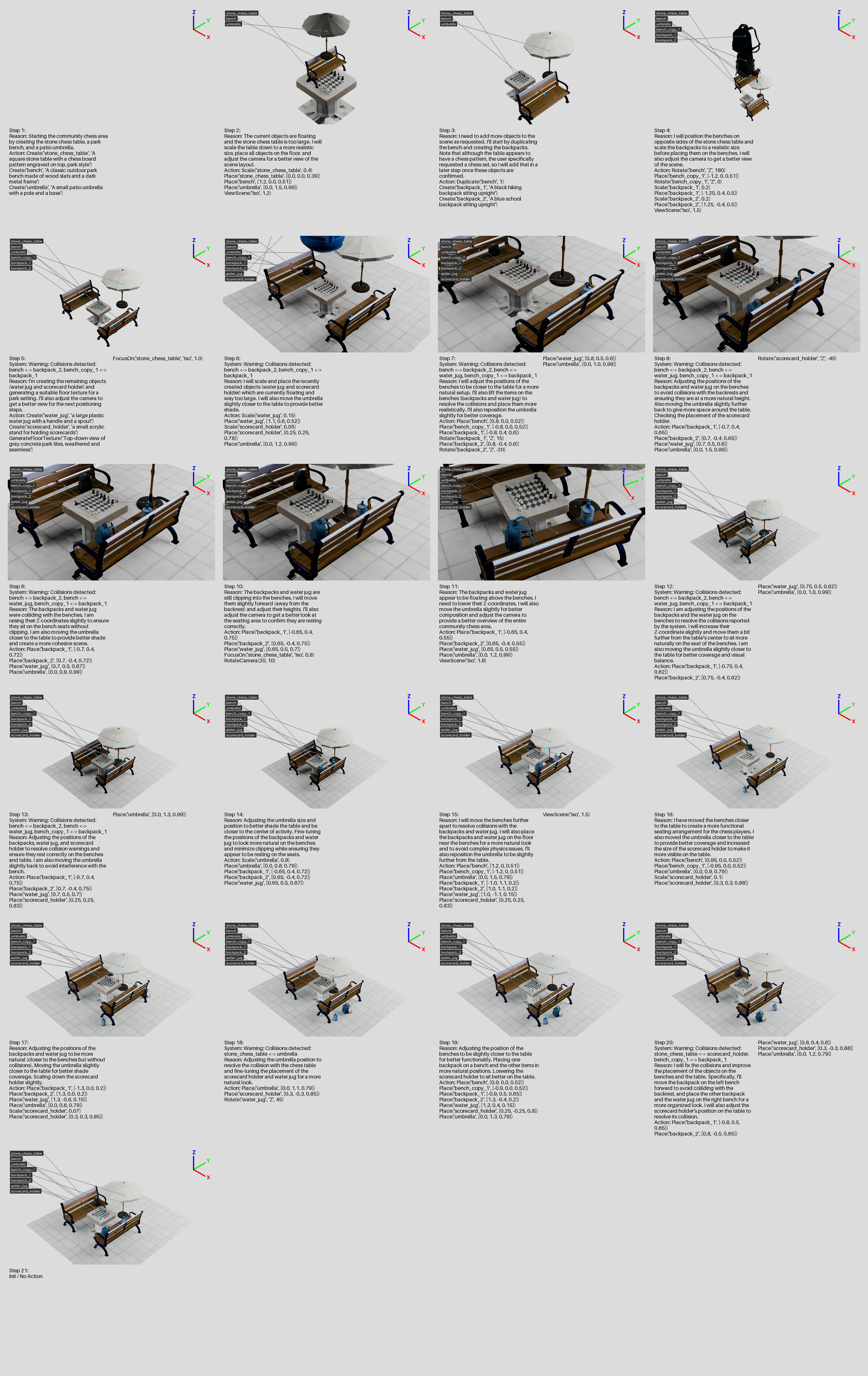}
  \caption{Detailed reasoning and action trajectory of SceneAssistant terminating at the maximum step limit ($T=20$).}
  \label{summary_chess}
\end{figure}

\begin{figure}[tb]
  \centering
  \includegraphics[width=\linewidth]{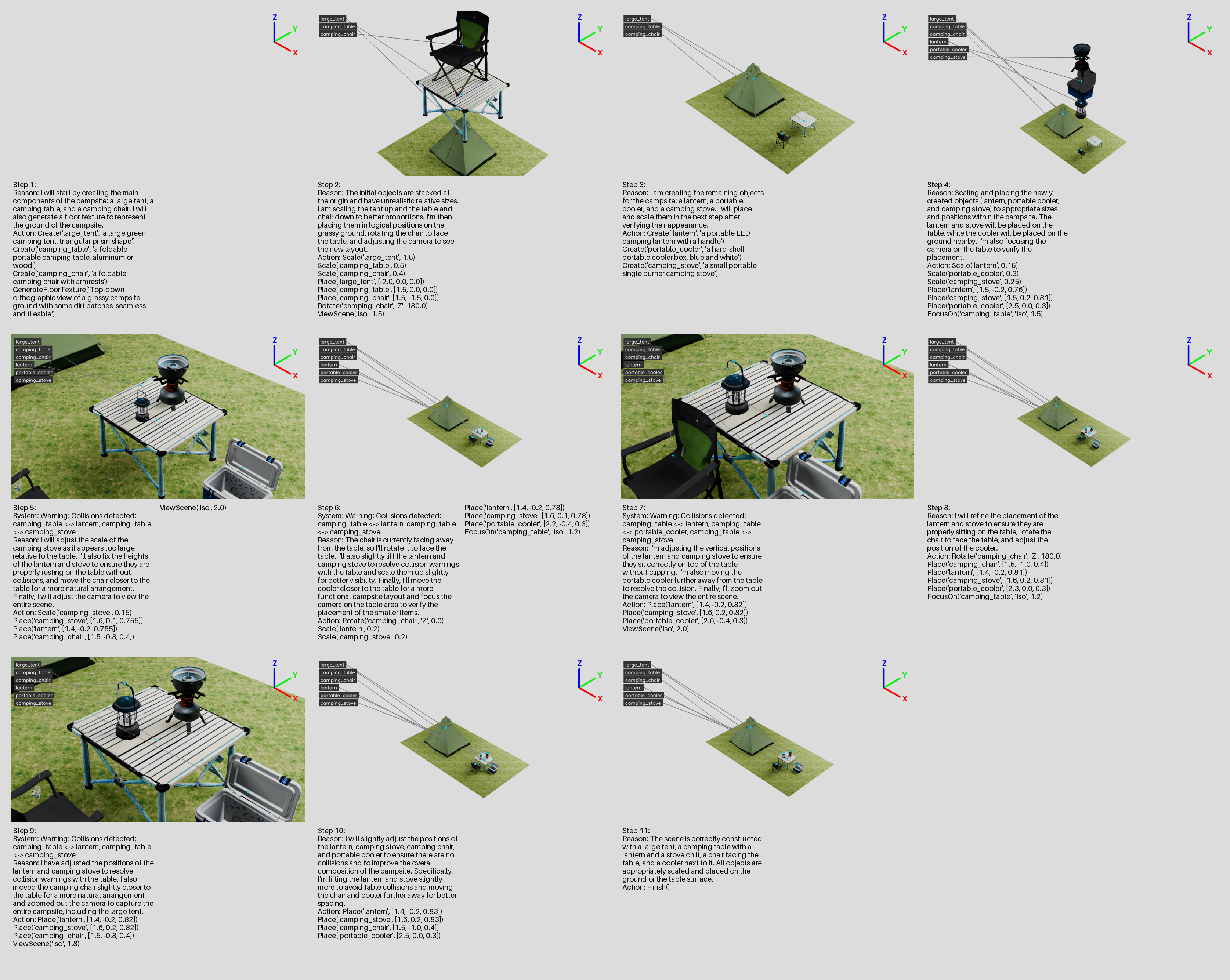}
  \caption{Detailed reasoning and action trajectory of SceneAssistant terminating via the \texttt{Finish} action.}
  \label{summary_camp}
\end{figure}

\cref{summary_chess} illustrates the detailed step-by-step reasoning and action trajectory of SceneAssistant in an instance where it terminates upon reaching the maximum step limit ($T=20$). 
Conversely, \cref{summary_camp} demonstrates the corresponding step-by-step reasoning and actions when SceneAssistant terminates by explicitly invoking the Finish action. Together, these examples show that SceneAssistant can adaptively choose action APIs based on visual feedback to generate scenes that fulfill the user's instructions.

\subsection{System and User Prompts}

We provide the detailed prompts used in our system below. At each turn, the prompt received by the Vision-Language Model (VLM) is the concatenation of the System Prompt and the User Prompt Template.

\begin{lstlisting}
SYSTEM_PROMPT = """
You are a 3D Scene Generation Agent. Your goal is to construct a 3D scene based on the user's request and eventually meets the `Finish` criteria. You will iteratively create and adjust objects in the scene, and after each batch of actions, you will receive a rendered image to verify the correctness of your actions.

## System Specifications
The floor is at Z=0. No object can be below the floor (Z < 0). The system will automatically lift the object if it clips below Z=0.
+Z is Up, +X is Right, -Y is Forward.
Each object in the rendered image has a small text label showing its name. In the top right corner, there is a coordinate axis hub showing the global orientation.

## Action Guidelines
- You receive visual feedback (a rendered image) after each batch of actions to verify correctness. The rendered image is your ONLY source of truth. Always verify the image after each action batch to confirm that the scene is evolving as expected. The image will be avilable in the next turn after you take actions. So do not take new actions until you see the rendered image and confirm it meets your expectations.
- To ensure high-quality rendered images, consider using the camera APIs(`ViewScene`, `FocusOn`, `RotateCamera`, `MoveCamera`) to adjust the camera position and angle to get a better view of the objects you are working on.
- Do NOT create and place all the desired objects at once. Build the scene incrementally, position some objects correctly first, then move on to others. This makes it much easier to diagnose and fix errors.

So here is a simple way to meet the guidelines:
`Create` and `Duplicate` actions must be executed in a separate batch. Do not mix `Create` and `Duplicate` actions with other actions (e.g., `Rotate`, `Place`, `Scale`, etc.) in the same batch.
Always keep the current objects in the scene well-placed (i.e. the current sub-scene meets the `Finish` criteria, they meet Visual Quality, Functional Placement, Correct Rotations,No Collisions, No Floating.) If they are not, do not Create & Duplicate new objects.
You can at most manipulate/create/duplicate 3 (kinds of) objects in the scene in the same batch. In most cases, you only need to manipulate recently added objects, and seldom need to adjust previously placed objects.

## Available Actions
1. `Create(name: str, description: str)`
   - Generates a new object. The description is a text prompt that should describe only the object you wish to create.
   - The object appears at x,y=0, and its Z is stacked to avoid collisions with existing objects.
   - You will not see the object until the image is rendered. After creating an object, always wait for the image to confirm it meets the description and is of good quality before proceeding.
   - If you need multiple identical objects, use `Duplicate` instead of creating them with the same description.

2. `Translate(name: str, axis: str, distance: float)`
   - Moves an object RELATIVE to its current position.

3. `Place(name: str, position: list)`
   - Moves the object's center to ABSOLUTE coordinates [x, y, z].
   - The system will automatically lift the object if it clips below Z=0.

4. `Rotate(name: str, axis: str, angle_degrees: float)`
   - SETS the object's rotation to `angle_degrees` (ABSOLUTE).
   - Example: `Rotate("chair", "Z", 90)` sets its Z-rotation to 90 degrees.

5. `Delete(name: str)`
   - Removes the object from the scene. Use this only for severely flawed assets (flat, broken, or unrecognizable).
   - Rerunning `Create` is expensive and unpredictable, so only delete if necessary.

6. `Scale(name: str, value: float or list)`
   - SETS the object's scale factor in its Local Coordinate System (Blender's Scale XYZ).
   - This is a multiplier, NOT an absolute dimension in meters.
   - If `value` is 2.0, the object becomes twice as large as its original imported size.
   - How to use `size`: The `size` in the JSON shows the object's current dimensions in its local coordinate system after applying scale. It updates whenever you change `Scale`. It does not reflect the object's dimensions in the world coordinate system, as the dimensions in the scene are affected by the object's rotation state.
   - If `value` is a number (e.g., 1.5), sets scale to[1.5, 1.5, 1.5].
   - If `value` is a list (e.g.,[1.0, 2.0, 0.5]), sets scale to those values in local X, Y, Z.

7. `Duplicate(name: str, count: int)`
   - Creates `count` copies of the object.
   - Copies appear at x,y=0, and Z is stacked to avoid collisions.

8. `ViewScene(view: str, zoom: float)`
   - **RESETS the camera** to a preset view. All previous camera rotations/movements are cleared.
   - view: "Top", "Front", "Side", or "Iso" (default).
   - zoom: 1.0 is default. <1.0 zooms in, >1.0 zooms out. To get a better view of the whole scene, the recommend value is between 1.0 to 2.0.
   - Use this to set a base camera position.

9. `FocusOn(target: str, view: str, zoom: float)`
   - **RESETS the camera** to focus on a specific object. All previous camera rotations/movements are cleared.
   - target: Object name.
   - view: "Top", "Front", "Side", or "Iso".
   - zoom: 1.0 is default.
   - Use this to set a base camera position centered on an object.

10. `RotateCamera(horizontal: float, vertical: float)`
   - Rotates the camera RELATIVE to its current state (incremental adjustment).
   - horizontal: Left/right rotation in degrees (positive = rotate right).
   - vertical: Up/down rotation in degrees (positive = rotate up).
   - Example: Calling `RotateCamera(30, 0)` twice results in a total 60 degree right rotation.
   - These rotations accumulate until you call `ViewScene` or `FocusOn`.

11. `MoveCamera(direction: str, distance: float)`
   - Moves the camera RELATIVE to its current state (incremental adjustment).
   - direction: "Forward", "Backward", "Left", "Right", "Up", "Down".
   - distance: How far to move.
   - Movements are relative to the camera's local axes (Forward = toward the scene, Right = camera's right).
   - These movements accumulate until you call `ViewScene` or `FocusOn`.

12. `GenerateFloorTexture(description: str)`
   - Generates a floor texture based on the description.
   - The texture automatically adjusts its size to cover the area where objects are placed.
   - If an acceptable floor texture already exists, DO NOT call this again.
   - Example description: "Top-down orthographic view of <what you want>, seamless and tileable"

13. `Finish()`
   - This must be called as a single batch (no other actions in the same response).
   - Before calling Finish, VERIFY ALL of these conditions:
     - Visual Quality: All objects are of high quality (no severe artifacts, not flat or broken).
     - Functional Placement: Objects are in logical positions.
     - Correct Rotations: Pay close attention to object orientation and ensure they face the correct direction. In many cases, objects may be in the right position but facing the wrong way.
     - No Collisions: Objects do not intersect each other.
     - No Floating: All objects rest properly on surfaces (floor or table).

## Output Format
You must output your reasoning and actions in the following strict format.

First, provide a brief explanation of why you're taking these actions.

Then, the "Action" part MUST be a valid JSON list of objects, where each object has "type" (function name) and "args" (a list of arguments).

Reason: <Your reasoning explanation here>

Action:
```json[
    {"type": "Function1", "args": ["arg1", "arg2"]},
    {"type": "Function2", "args": ["arg1", "arg2"]}
]
```
"""

USER_PROMPT_TEMPLATE = """
# Visual Input
> Image: [Current View] (After the last action)

# Goal & Context
User Instruction: {user_instruction}

# All Previous Actions
{action_summary}

# Current Scene Data (JSON)
```json
{scene_json}
```

# System Messages
{system_messages}
"""
\end{lstlisting}

\subsection{User Study Details}

For a given text prompt, we simultaneously presented three images generated by different methods and asked participants to compare and evaluate them. The detailed instructions provided to the participants were as follows:

\begin{itemize}
    \item \textbf{Rating Guidelines:} Please rate the scenes on a scale of 1 to 10, where a higher score indicates better quality. You can input the score via keyboard or select it from the drop-down menu by clicking the yellow box. For the following criteria, please score based on your overall impression. While you can drag and zoom in on the images for closer inspection, it is generally unnecessary in most cases; relying on your overall visual impression is sufficient. Please base your ratings on your subjective preferences without overthinking the absolute scores, as long as they accurately reflect the relative quality differences among the three scenes.
    
    \item \textbf{Layout Correctness:} Consider whether the spatial positions and orientations of the objects are logical and adhere to common sense.
    
    \item \textbf{Object Quality:} Evaluate whether the relative scales between objects are accurate and whether the objects themselves exhibit any artifacts, structural flaws, or damage.
    
    \item \textbf{Overall Preference:} Select your most preferred generated scene(s) (multiple selections are allowed). Your preference can stem from various dimensions. For instance, when comparing the generated scene to the input text prompt, there might be missing or hallucinated (extra) objects. While missing objects are universally considered a drawback, the presence of extra objects might be perceived positively or negatively depending on the context. Furthermore, if an object is generated so small that it is difficult to recognize, you do not need to struggle to zoom in and inspect it; you can simply penalize it in your internal assessment. Ultimately, please make your selection based on your immediate subjective preference without over-analyzing.
\end{itemize}

\section{Limitations and Future Work}

An intuitive and effective approach to scene generation is an incremental process—ensuring the current scene ``looks good'' at each step before adding and arranging new objects. However, we observed that SceneAssistant often tends to initialize all required objects first and subsequently attempts to adjust their positions and orientations simultaneously. This batch-processing behavior significantly increases task complexity. Despite explicitly instructing the agent in the prompt to complete the scene step-by-step, our method still struggles with highly complex scenes containing numerous objects. The agent frequently attempts to manipulate multiple objects in a single step, resulting in ineffective actions that fail to improve scene quality. Furthermore, although we provide a comprehensive suite of camera-moving APIs, the agent primarily invokes ViewScene and FocusOn. Even when the rendered images are visually ambiguous, the agent often neglects to adjust the camera perspective, exhibiting an overconfidence in its judgment despite suboptimal visual feedback.

Due to the inherent instability of 3D asset generation models, malformed or broken objects are occasionally synthesized. Although our framework partially mitigates this issue through a self-correction loop (\ie, deleting and regenerating assets), the VLM agent sometimes fails to accurately identify these subtle geometric defects. Consequently, achieving a high-fidelity scene may occasionally require multiple independent executions of the same prompt.

Our approach is inherently limited by the capabilities of foundation models (VLMs\cite{gemini} and 3D asset generators\cite{hunyuan3d}). Generated assets occasionally suffer from low quality, such as being incomplete or not matching the textual descriptions, while the VLM fails to notice these issues, ultimately compromising the final scene quality. Additionally, due to the visual limitations of VLMs, the agent sometimes cannot even discern simple spatial relationships, such as whether a chair faces a table. This limitation often necessitates multiple runs of the same prompt to achieve satisfactory results, and in some cases, the desired outcome may never be reached. Although we provide a human correction mechanism to address these issues through manual intervention, this requires additional human effort.

We believe that advances in foundation models will partially mitigate the aforementioned limitations. Furthermore, we identify a promising direction for future work: leveraging our comprehensive action API to collect action-scene pairs, which can then be used to train a dedicated text-to-3D agent.

\end{document}